\documentclass[lettersize,journal]{IEEEtran}
\IEEEoverridecommandlockouts

\usepackage{amssymb}
\usepackage[cmex10]{amsmath}
\usepackage{stfloats}
\usepackage{graphicx}
\usepackage{subfigure}
\usepackage{tabularx}
\usepackage{epsfig,epsf,color,balance,cite}
\usepackage{verbatim}
\usepackage{url}
\usepackage{bm}
\usepackage{booktabs}
\usepackage{siunitx}

\usepackage{amsmath}
\allowdisplaybreaks[2]
\usepackage{ntheorem}

\usepackage{algorithm}
\usepackage{algorithmic}

\usepackage{color}
\definecolor{myc1}{rgb}{0,0,0}

\begin{document}

\title{Semantic-Aware Visual Information\\ Transmission With Key Information Extraction Over Wireless Networks}

\author{Chen Zhu, Kang Liang, Jianrong Bao, Zhouxiang Zhao, Zhaohui Yang, Zhaoyang Zhang,\\ and Mohammad Shikh-Bahaei
\thanks{C. Zhu is with School of Communication Engineering, Hangzhou Dianzi University, Hangzhou 310018, China, and also with Polytechnic Institute, Zhejiang University, Hangzhou 310015, China (e-mail: zhuc@zju.edu.cn).}
\thanks{K. Liang, Z. Zhao, Z. Yang, and Z. Zhang are with College of Information Science and Electronic Engineering, Zhejiang University, Hangzhou 310027, China (e-mails: \{3220104354, zhouxiangzhao, yang\_zhaohui, ning\_ming\}@zju.edu.cn).}
\thanks{J. Bao is with School of Communication Engineering, Hangzhou Dianzi University, Hangzhou 310018, China (e-mail: baojr@hdu.edu.cn).}
\thanks{M. Shikh-Bahaei is with Department of Engineering, King’s College London, London, UK (e-mail:m.sbahaei@kcl.ac.uk).}
}

\maketitle

\begin{abstract}
The advent of 6G networks demands unprecedented levels of intelligence, adaptability, and efficiency to address challenges such as ultra-high-speed data transmission, ultra-low latency, and massive connectivity in dynamic environments. Traditional wireless image transmission frameworks, reliant on static configurations and isolated source-channel coding, struggle to balance computational efficiency, robustness, and quality under fluctuating channel conditions. To bridge this gap, this paper proposes an AI-native deep joint source-channel coding (JSCC) framework tailored for resource-constrained 6G networks. Our approach integrates key information extraction and adaptive background synthesis to enable intelligent, semantic-aware transmission. Leveraging AI-driven tools, Mediapipe for human pose detection and Rembg for background removal, the model dynamically isolates foreground features and matches backgrounds from a pre-trained library, reducing data payloads while preserving visual fidelity. Experimental results demonstrate significant improvements in peak signal-to-noise ratio (PSNR) compared with traditional JSCC method, especially under low-SNR conditions. This approach offers a practical solution for multimedia services in resource-constrained mobile communications.
\end{abstract}


\IEEEpeerreviewmaketitle

\section{Introduction}
The advent of wireless communication technology has precipitated a paradigm shift in the realm of high-resolution image and video transmission, with profound implications for diverse domains such as mobile communication, multimedia services, the Internet of Things (IoT), and virtual reality \cite{ref34,ref35,ref36,10550151}. Nevertheless, as the demand for real-time and high-quality transmission continues to escalate, the discrepancy between the ever-increasing volume of data and the constrained capacity of the channel is intensifying \cite{ref20,ref21,ref22,e26050394}. This challenge is further exacerbated by the complex and dynamic nature of wireless channels, which are subject to bandwidth limitations, noise interference, energy loss, and other factors \cite{ref37,ref38,ZHAO2024107055}. These issues make it difficult to simultaneously meet the requirements of high-quality transmission, low latency, noise resistance, and low power consumption, especially in the context of emerging 6G networks that aim to deliver unprecedented levels of performance characterized by ultra-high speeds, minimal latency, and massive connectivity.

Traditional image transmission schemes, based on Shannon’s separation theorem \cite{ref1}, typically rely on independent source coding (e.g., JPEG, H.264) and channel coding (e.g., LDPC, Turbo codes). While this cascading approach exhibits certain theoretical optimality, it encounters substantial challenges in practical implementations. On one hand, source coding is incapable of perceiving channel conditions, which results in an inability to flexibly adjust when channel conditions deteriorate. On the other hand, channel coding struggles to prioritize the protection of important information from the source, which may lead to decreased transmission efficiency. These limitations highlight the need for a more adaptive and intelligent approach to image transmission, especially in the context of 6G networks that demand dynamic optimization and intelligent management.

In order to overcome these challenges, researchers have tried a variety of methods in recent years, such as semantic communication \cite{ref23,ref24,ref25,ref26,ref27,ref28,10915662} and generative AI for communication \cite{ref29,ref30,ref31,ref32,ref33} to reduce the transmission pressure and improve the transmission efficiency. In addition, Deep Joint Source Channel Coding (JSCC) transmission is a very effective strategy in reducing transmission pressure and improving transmission quality. Significant gains can be achieved using JSCC in various schemes such as vector quantization and index assignment 
\cite{ref2,ref3,ref4,ref48,ref49,ref50,10734747}. In recent years, deep learning has been applied to a variety of fields due to its significant effect, as well as to the field of communication \cite{ref39,ref40,ref41,ref42,ref43,ref44,ref45,ref46,ref47,11006980}. Due to its superior performance in extracting various complex features, deep learning has also been extended to the JSCC model for wireless image transmission \cite{ref5,ref6,ref7,ref8,ref9,ref10,ref11,ref12,ref13}. This technology directly encodes image features into transferable signals through end-to-end deep learning models, achieving joint optimization of source coding and channel coding. In these papers, they all proposed the deep learning-based JSCC model to extract image features at the receiver through joint source-channel coding, thus greatly reducing the amount of information incoming to the channel. Then the received information is used to restore the image through the trained deep learning model to obtain the final result. Compared to the traditional information transmission model, this method can greatly reduce the occupation of channel resources while ensuring high image transmission quality. In this paper, we adopt an adaptive rate deep JSCC model \cite{ref13}. This model can significantly improve image transmission quality under low signal-to-noise ratio (SNR) conditions and effectively reduce frequency band utilization by dynamically adjusting transmission rates in high SNR or simple image content scenes. This approach improves the overall efficiency of wireless communication systems.

However, although the model performs well in many application scenarios, there are still some limitations. For example, for high-resolution images, their computational complexity and demand for transmission resources are still high, resulting in high computational resource requirements and low transmission efficiency. In addition, in images with complex backgrounds, the model's ability to transmit and reconstruct specific target areas may be limited, thereby affecting the overall visual quality of the image. For high-resolution image transmission, there are issues of high computational power requirements and high consumption of computing resources. It requires a GPU with large graphics memory to run the model and also consumes high computing resources for model inference.

To address this challenge, the current JSCC model-based scheme for wireless transmission of high-resolution images effectively mitigates the issue of the deep JSCC model's requirement for high-performance GPUs and substantial graphic memory \cite{ref14}, as well as the deep JSCC model's significant consumption of computational resources \cite{ref15}. In this context, we propose a method for JSCC transmission of high-resolution images which involves extracting essential information and leveraging an image information database to compress the incoming image data. This approach not only reduces the need for computational power and resource consumption but also sustains high image quality with the assistance of the database. The fundamental concept of this methodology entails initially extracting key information from high-resolution images, dissociating the key information from the image background, and constructing a resource library at both the transmitting and receiving ends. Consequently, during JSCC model transmission of high-resolution images, only the extracted feature components are transmitted, and the final image is synthesized at the receiving end via background matching, thereby alleviating transmission and computational pressures while enhancing image quality.

By integrating key information extraction and background matching within the Deep JSCC framework, the proposed method effectively diminishes computational complexity and resource demands pertinent to high-resolution image transmission, thereby augmenting the efficiency and robustness of wireless communication. This dynamic optimization of transmission rates and image resolution is congruent with the objectives of 6G networks, which seek ultra-high speed, minimal latency, and extensive connectivity. Furthermore, by reducing the dependency on high-performance GPUs and substantial memory capacities, this solution facilitates the implementation of intelligent and adaptive transmission systems in resource-constrained environments. Overall, this work establishes an approach for efficient and low-latency image and video transmission in 6G networks, propelling the evolution of next-generation wireless technologies.

We provide a list of notation used in our paper in Table \ref{table1} for reference.

\begin{table}[ht]
\scriptsize
\centering
\caption{Notations}\label{table1}
\begin{tabular}{|c|l|}
\hline
\textbf{Notation} & \textbf{Description} \\ \hline
$P$ & Key points in the human body \\ \hline
$H$ & Convex hull of human body contour \\ \hline
$M(x, y)$ & Mask function (0: shielded, 1: reserved) \\ \hline
$X_m(x, y)$ & Masked image \\ \hline
$B_m(x, y)$ & Masked background \\ \hline
$D_X$ & Descriptors from $X_m$ \\ \hline
$D_B$ & Descriptors from $B_m$ \\ \hline
$Match(D_X, D_B)$ & Matching function based on Hamming distance \\ \hline
$d_{X_i}$ & Descriptor of $X_m$ \\ \hline
$d_{B_i}$ & Descriptor of $B_m$ \\ \hline
$H(d_{X_i}, d_{B_i})$ & Hamming distance between descriptors \\ \hline
$T$ & Threshold for matching \\ \hline
$N_k$ & Number of matched feature pairs for $B_k$ \\ \hline
$B^*$ & Best-matching background \\ \hline
$N_{min}$ & Minimum match threshold \\ \hline
$X_E$ & Extracted key image information \\ \hline
$(x_1, y_1)$ & Upper-left corner of character area \\ \hline
$(x_2, y_2)$ & Lower-right corner of character area \\ \hline
$X_s$ & Latent features from source encoder $E_s$ \\ \hline
$C$ & Channels in input image $X$ ($C = 3$) \\ \hline
$C_s$ & Channels in $X_s$ ($C_s = 256$) \\ \hline
$H_s$ & Height of $X_s$ ($H_s = H / 4$) \\ \hline
$W_s$ & Width of $X_s$ ($W_s = W / 4$) \\ \hline
$X_c'$ & Output of channel encoder $E_c$ \\ \hline
$C_c'$ & Channels in $X_c'$ ($C_c' = 16$) \\ \hline
$H_c$ & Height of $X_c'$ ($H_c = H_s$) \\ \hline
$W_c$ & Width of $X_c'$ ($W_c = W_s$) \\ \hline
$W$ & Binary mask for active features \\ \hline
$X_c$ & Compressed features after applying $W$ \\ \hline
$C_c$ & Channels in $X_c$ ($C_c = 8$) \\ \hline
$L$ & Length of compressed feature vector ($L = 2 \times H_c \times W_c$) \\ \hline
$\mu$ & Compression ratio ($\mu = \frac{H(X_c)}{H(0)} = 1/3$) \\ \hline
$x_n$ & Transmission decision for user $n$ ($x_n \in \{-1, 0, 1\}$) \\ \hline
$\rho_n$ & Communication quality for user $n$ \\ \hline
$\alpha$ & Quality factor for key information transmission ($0 < \alpha < 1$) \\ \hline
$d_n$ & Information transmitted by user $n$ \\ \hline
$D_n$ & Information for direct transmission \\ \hline
$C_n$ & Information for key information transmission ($C_n < D_n$) \\ \hline
$N_n$ & Noise in user $n$'s channel \\ \hline
$T$ & Maximum transmission delay \\ \hline
$P_{max}$ & Total power at base station \\ \hline
$r_n$ & Transmission rate for user $n$ (Shannon's formula) \\ \hline
$G$ & System gain \\ \hline
$L$ & Power loss during transmission \\ \hline
$P_n$ & Power consumption for user $n$ \\ \hline
$P_{n0}$ & Power for direct transmission \\ \hline
$P_{n1}$ & Power for key information transmission \\ \hline
\end{tabular}
\end{table}

\section{Key Information Extraction and Background Synthesis}
\begin{figure*}[ht]
    \centering
    \includegraphics[width=0.9\linewidth]{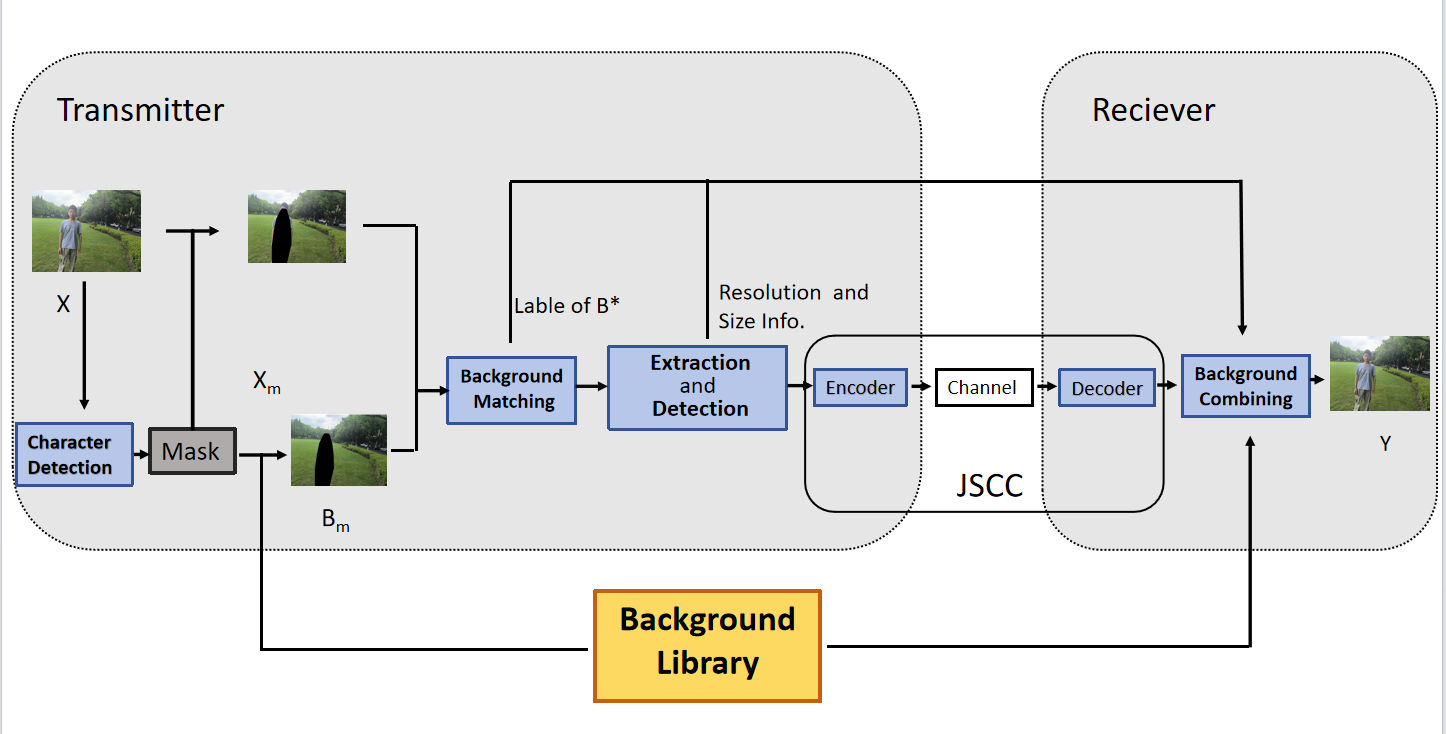}
    \caption{Structure of Our Key Information Extraction and Background Synthesis Model.}
    \label{fig:enter-label}
\end{figure*}

\subsection{Motivation}
In some specific use cases, such as video streaming, surveillance, or remote conferencing, the background may remain the same for a certain number of images or frames, or we may not need to recover the background with high precision. In these scenarios, rather than transmitting the entire image with its detailed background, we can take advantage of a background library. This library stores a collection of common background images that can be reused for matching, which significantly reduces the need to transmit redundant background data with every frame.

By establishing such a background library, we can compare the current image with the stored backgrounds and identify the best-matching background, rather than encoding and transmitting the background portion of the image each time. This approach allows us to isolate and transmit only the foreground information, i.e. the subject of interest, which is the main focus of most applications like object detection, human pose estimation, and feature extraction.

This method is particularly advantageous when the background is static or changes slowly over time, as it eliminates the need for recomputing or transmitting large portions of the image that do not change. Therefore, the essential communication overhead can be greatly reduced, leading to more efficient bandwidth usage and faster transmission times. Additionally, by reducing the size of the transmitted image, we can allocate more resources to enhancing the quality of the foreground data, improving the overall transmission efficiency, and reducing the impact of background noise on feature extraction.

At the same time, the method uses the deep JSCC model to transmit the extracted key information, so that only a small part of the information in the high-resolution image enters the deep JSCC model, which greatly reduces the computing resources and computing power required for the model transmission. This enables it to transmit higher resolution images under lower conditions, so that the resulting image quality is higher.

In short, while the background is relatively stable, using the background database enables us to save bandwidth and computing resources while still maintaining high-quality foreground data. And it can reduce the threshold of using this method, that is, the requirement of the deep model for high-performance GPU and large video memory. This method is very important for optimizing the performance of real-time image transmission system when the computing conditions and transmission conditions are relatively limited.
\subsection{Background Matching}

Before transmission, we set up a background library to match the background of the images to be transmitted. If background matching is successful, we proceed with subsequent feature extraction to improve image quality. If the matching fails, the image is transmitted directly through the JSCC model after resolution detection and compression.

The Mediapipe library \cite{ref16,ref17} is a powerful open source framework developed by Google for building multi-modal, cross-platform machine learning solutions. It provides a collection of pre-trained models and tools for real-time tracking and detection of human body landmarks, including face, hand, and full-body pose estimation. Mediapipe’s pipeline is designed for efficiency, enabling accurate detection even on resource-constrained devices.We use the Mediapipe library to detect key points in the body of people $\mathit{P} =\{(x_i,y_i)|i=1,2,3...,n\} $ in image X and  generate convex hull $\mathit{H}$  containing human body contour. Then we build the mask $\mathit{M}$
$$ 
 \mathit{M(x,y)} = \left \{ 
 \begin{array}{rcl}
 0 ,& if(x,y) \in H, \\
 1,& otherwise, \\
 \end{array}
\right .
$$
where $M(x, y) = 1$ indicates the area to be reserved, and $M(x, y)=0$ indicates the area to be shielded.This mask will be added to the image $\mathit{X}$ and the background $\mathit{B}$ to get $\mathit{X_m}$ and $\mathit{B_m}$ to prevent the interference caused by the characters in the image to the background matching.
$$ 
 \mathit{X_m(x,y)} = \left \{ 
 \begin{array}{rcl}
 X(x,y) ,& if M(x,y) = 0, \\
 0,& if M(x,y) = 1. \\
 \end{array}
\right .
$$
$$ 
 \mathit{B_m(x,y)} = \left \{ 
 \begin{array}{rcl}
 B(x,y) ,& if M(x,y) = 0,\\
 0,& if M(x,y) = 1. \\
 \end{array}
\right .
$$
After obtaining the masked images $X_m$ and $B_m$ , the ORB (Oriented FAST and Rotated BRIEF) \cite{ref18} algorithm is applied to extract feature points and compute descriptors for both the masked image and the candidate backgrounds in the background library. Let the feature descriptors extracted from  $X_m$ and $B_m$ be represented as $D_X$ and $D_B$ , respectively.The descriptors are matched using the Hamming distance to identify corresponding feature points between $X_m$ and $B_m$ . The matching process can be described mathematically as follows:
$$Match(D_X,D_B) = \{(d_{X_i},d_{B_j}) | H(d_{X_i},d_{B_j}) \leq T \},$$
where $d_{X_i}$ and $d_{B_j}$ are the feature descriptors of $X_m$ and $B_m$ , respectively. $H(d_{X_i},d_{B_j})$ is the Hamming distance between two binary descriptors. $T$ is a predefined threshold for determining a match.For each background $B_k$ in the background library, the number of matched feature pairs is calculated: $N_k =  |Match(D_X,D_{B_k})|$.The best-matching background is identified as the one with the maximum number of matches: $B^* = \mathop{argmax}\limits_{k}N_k$,where $B^*$ is the background selected as the best match for the image X.If no background $B_k$ satisfies a minimum match threshold $N_{min}$ , the background matching is considered unsuccessful. In this case, the image X is transmitted directly using the JSCC model after resolution detection and compression.
\subsection{Key information extraction}
After successfully matching the background, we extract the characters from the successfully matched images. For this task, we use the Rembg library, which is an efficient and easy-to-use tool for background removal in images. Rembg utilizes deep learning models trained to segment objects (or characters) from their backgrounds, providing high-quality foreground masks without the need for manual annotation. The library is designed to automatically detect and remove the background from images, leaving only the foreground (the subject of interest). It supports multiple types of segmentation tasks, such as human body, products, and objects, and provides an API that can be integrated easily into image processing pipelines.
The library works by applying pre-trained neural networks that segment the foreground and background based on pixel-level classifications, making it both fast and accurate. This process ensures that we can isolate the character from the image and remove any unwanted background interference. Rembg also supports various image formats and works with high precision on both structured and unstructured backgrounds, making it a reliable choice for foreground extraction.
Through this function library, we can accurately identify the character in image $\mathit{X}$, and use the coordinates of two points $(x_1,y_1)$ and $(x_2,y_2)$ to represent the coordinates of the upper left corner and lower right corner of the area where the character is located. Then we can extract the key image information $\mathit{X_E}$ from the original image $\mathit{X}$ according to this coordinate.
$$X_E=X[x_1:y_1,x_2:y_2].$$
At the same time, we record position and size information during extracting to facilitate subsequent image restoration. After extracting, we obtained the character features and images with minimal background information. The main purpose of this operation is to reduce the channel resource occupation of the image background when using the JSCC model, as well as the influence of the image background on the character features when generating the image, in order to improve the image quality of the character features after transmission. After completing the extraction operation, it can be transmitted through the JSCC model.

\subsection{Resolution detection and compression}
Thanks to the development of photography technology today, the resolution of many images has been greatly improved. To cope with situations where the image resolution is too high or the computing power is low in practical applications, we need to add a step of resolution detection compression after the above process to ensure that the image resolution entering the JSCC model is not too high. So we set a suitable resolution threshold  that matches the computing power. When the resolution of the image to be input into the JSCC model exceeds the threshold, we will compress the image resolution proportionally to make it just below the threshold for subsequent steps.
\subsection{Transmission with JSCC Model}
\begin{figure}[ht]
    \centering
    \includegraphics[width=1\linewidth]{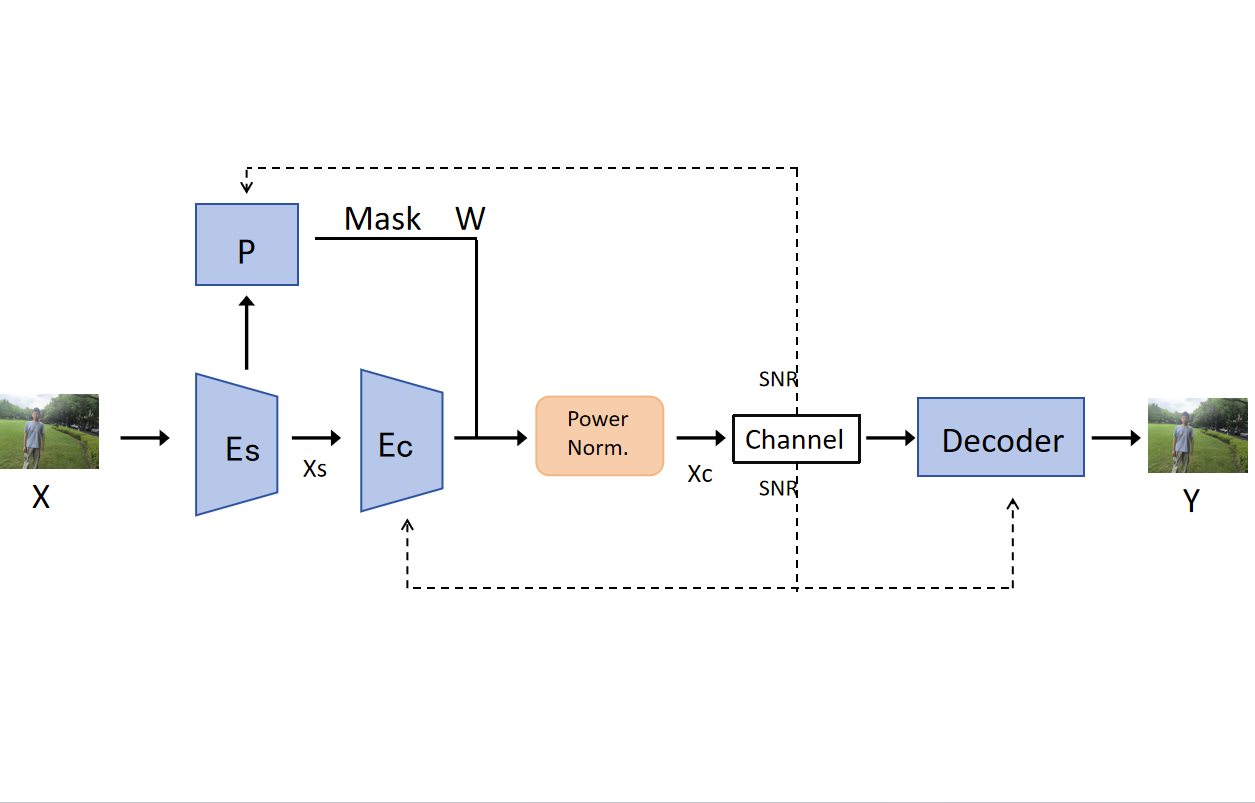}
    \caption{Structure of the considered JSCC model.}
    \label{fig:1}
\end{figure}

The structure of the considered model is shown in Figure~\ref{fig:1}. In this model, the image $X \in \mathbb{C}^{C \times H \times W}$ is first passed through the source encoder $E_s$, which includes one convolution layer and two down-sampling layers, to extract the latent $X_s \in \mathbb{C}^{C_s \times H_s \times W_s}$ from the source image $X$. 

Then $X_s$ passes through the channel encoder $E_c$ and generates $X_c' \in \mathbb{C}^{C_c' \times H_c \times W_c}$, where $C_c'=16$, $H_c=H_s$, and $W_c=W_s$. At the same time, $X_s$ will pass through a policy network to generate a binary mask $W$ to select active features of $X$. The mask $W$ will convert $X_c'$ to $X_c \in \mathbb{C}^{C_c \times L}$
The deep JSCC model can perform deep sampling and feature extraction of the image at the receiver, so as to greatly reduce the occupation of channel resources when transmitting images. Moreover, this model also has great advantages in lightweight. The model only employs a single-layer neural network, which enables it to be used and trained with low demand. In addition, when training the model, the SNR of the channel will be added to the trained model, so that it can adaptively adjust according to the real-time SNR of the channel, so that we do not need to train the SNR separately in order to obtain the optimal result under a certain SNR. Moreover, this model also has great advantages in the quality of image transmission. Considering the above points, we adopt the deep JSCC model as our experimental model to carry out subsequent experiments.
\subsection{Image restoration and synthesis}
After obtaining the image transmitted by the JSCC model, we will restore the image to its original size based on the cropped data recorded during previous feature extraction. After restoring to the corresponding size, we will utilize the background synthesis function in the Rembg library to combine the extracted and transmitted features with the background in the previously matched background library, thus completing the entire feature extraction transmission and background synthesis operation.

\section{Adaptive Dynamic Background Library}
\subsection{Motivation}
In the method proposed above, we can rely on the shared independent background database and JSCC transmission model to compress the amount of image information to be transmitted in a large scale. In today's demand for visual information transmission, in many cases, the demand for video transmission is far greater than that for image transmission, and the above methods are more suitable for video image transmission. However, in addition to some special video transmission requirements, the establishment of background database is often very difficult. Because the background of the video that people transmit most of the time is not certain and predictable, it is difficult for people to predict what background the future video transmission will be under, so as to build a corresponding background database for it. Therefore, on the basis of the above JSCC model based on the background database, we further build an adaptive dynamic background database model, which broadens the use scenarios of the above model and makes it more suitable for modern people's application scenarios for video transmission.
\subsection{Video Background Recognition and Dynamic Background Database Construction}
For video scenes with relatively fixed background, we can directly collect enough background information at the receiver. Although the background will be blocked to a certain extent due to the existence of the foreground, in fact, the background information we need has completely appeared in the original video, so we only need to extract and splice it to get the final background we need, and then send it to the receiver as the background database. First, we obtain video frames $\mathbb{X}$ = $\{X_1,X_2,X_3,...,X_n\}$($X \in \mathbb{C}^{C \times H \times W}$), and then use Rembg library to separate foreground and background, so as to obtain the corresponding mask $\mathbb{M} = \{M_1,M_2,M_3,...,M_n \}$($M \in \mathbb{C}^{ H \times W}$, $[M_n]_{ij}={0,1}$,where 0 for foreground 1 for background) of each video frame.Then we perform an AND operation on the mask value of each pixel to get the final mask $M_m$ with the largest background information.However, each pixel of each video frame will be supplemented according to the mask to obtain the final background $X_B$.As shown in Figure 3, we can get the final background by summing the background information in the video frame. Although sometimes the background of a certain part is always blocked, which leads to the incomplete background we finally get, it is also because the background of this area is always blocked, which means that the background information of this part is invalid. We can directly discard this part of the background and transmit the final background to the receiver as the database background.
\begin{figure}[ht]
    \centering
    \includegraphics[width=1\linewidth]{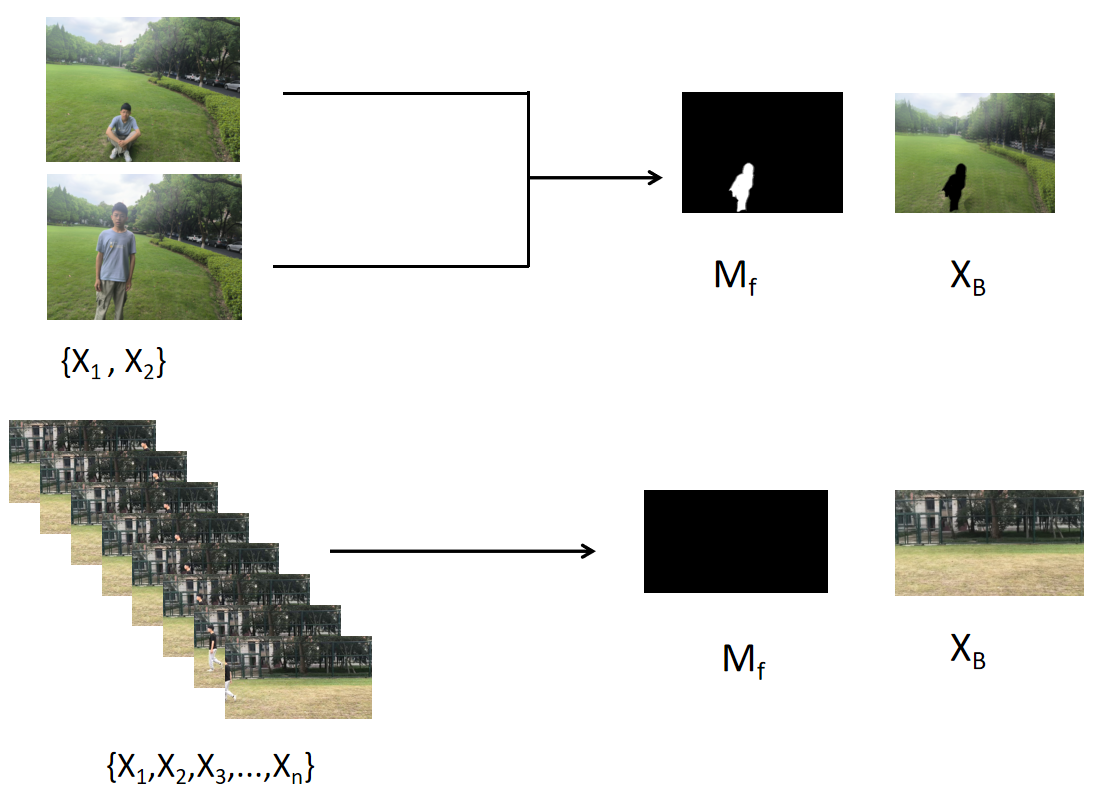}
    \caption{Get the maximum background information by integrating the background information of each video frame}
    \label{fig:3}
\end{figure}
\subsection{Processing of Dynamic Background }
A very obvious point is that the above method of building dynamic background database can only have obvious and high-quality results in application scenarios with relatively fixed background. However, in actual applications, the application scenarios we need are much more complex than those with a single fixed background, and the outstanding ones are the actual situations of dynamic background and real-time communication. The existence of dynamic background makes the image background of the video we need to transmit in a changing state, which makes us mix the background information of several other areas when processing the background of one area of the image, making the background information become confused and distorted. For the above situation, we choose the method of image splitting to construct multiple backgrounds to deal with the transmission of dynamic background video. Although the video background is dynamic, in general, a certain segment of the background in the video will stay in a relatively fixed area, so we can consider the video in this period of time to be in the same background, so we can split a dynamic background video into multiple static background videos, and then use the above method to build the background database. As for how to split the image, we use the ORB algorithm used in background matching. Because ORB algorithm recognizes the image similarity by means of each feature in the image, the influence of background difference caused by small range movement of perspective can be reduced to a certain extent. Through this algorithm, we can set a threshold of similarity, segment the video whose background similarity is within the threshold range for a period of time into a segment, and transmit the segment as a static background using the above method. Although this may increase the pressure of transmission, compared to direct transmission, it will save a lot of transmission resources.

\subsection{Problems in Real-time Video Communication}
In the case of real-time communication, due to the need to consider the dual effects of real-time and delay, on the one hand, we can not obtain all the effective background information from the video information to build a dynamic background database; on the other hand, if the computing resources are relatively tight, we can hardly guarantee low delay and high communication efficiency. Since the non real-time video communication is carried out under the condition that all information is fixed, we can easily establish the dynamic background database at the receiver. Then, in the range of delayed operation, the real-time video communication we can obtain is limited, and how the information in the subsequent video changes (such as whether the background changes, whether the foreground position changes) is unknown, which makes it difficult for us to obtain an appropriate background from the existing information and add it to the background database. However, although it is difficult for our method to be applied to real-time video communication at present, in view of the rapid development and wide application of AI, we have conceived a method to solve existing problems in the future by using generative AI. We imagine to use the above background synthesis method to synthesize the background of the first few frames of real-time communication, and then complete the missing parts through generative AI. Considering the randomness of generative AI, we will continue to synthesize the background, update the background database in real time, and restore the background information to the maximum extent in the subsequent communication process. Although this method will increase the amount of information in the incoming channel to a certain extent, for large video frames, the real-time background of intermittent transmission only occupies a small part of the transmission information, so there is no need to worry about the increase of communication burden caused by this method.

\subsection{Frame Generation of Video}
When considering how to reduce the pressure of video transmission, we once imagined that we could sample video frames to a certain extent at the input end, that is, only transmit some video frames according to the sampling rate, and the remaining video frames will be recombined into video after being generated by the generative frame interpolation model at the receiver. Then, if operated directly, it will lead to large consumption of computing resources and high transmission delay, which is very fatal for real-time video communication. Therefore, we hope to find some lightweight methods to generate video frames. As we mentioned earlier, we separate the key information in the video from the background, and only transmit the key information part to reduce the transmission pressure. In essence, this approach saves transmission resources by reducing the amount of image information. Similarly, this method can still be used in the video frame generation. By reducing the amount of information in the input image, the computational resources required for frame generation can be reduced as much as possible. So we used the model in \cite{ref19} to generate video frames. Because it is extremely lightweight, it can reduce the time delay while completing frame generation. But in this process, we also encountered other problems, because for the foreground of video time changes, it is difficult for a lightweight generative model to compare and complete the frame generation of some complex actions, and it can only have a better performance when the foreground is close to translation. This is a trade-off for our project, because if we want to achieve low latency and good frame generation performance for various application scenarios at the same time, we must adopt a more complex and large model, which is contrary to our original intention. So we still decided to adopt this model, but we gave some choices in the application, that is, the user can decide whether to adopt it or not. If the sudden extreme situation occurs, the user can use this method to improve the communication quality as much as possible.

\section{Adaptive Base Station Transmission}
Considering how to select the appropriate transmission scheme according to the real-time channel conditions and the base station conditions in practical applications (i.e., direct transmission and resynthesis transmission through key information extraction), we propose an adaptive base station transmission scheme. By obtaining the current channel conditions, user transmission requirements and the total power that the base station can use for wireless communication in real time, the base station provides the most appropriate transmission scheme for the current situation.Assuming that there are N users who need wireless transmission at this time, let $x_n\in \{-1, 0, 1\}$ respectively indicate that the nth user does not transmit, transmits directly and transmits with the above method. The user's communication quality $\rho_n$ :
$$ 
 \rho_n = \left \{ 
 \begin{array}{rcl}
 0 ,& x_n =-1, \\
 1 ,& x_n =0, \\
 \alpha ,& x_n =1,
 \end{array}
\right .
$$
where $0<\alpha<1$.At this time, the amount of information transmitted by the user is $d_n$:
$$ 
 d_n = \left \{ 
 \begin{array}{rcl}
 0 ,& x_n =-1, \\
 D_n ,& x_n =0, \\
 C_n ,& x_n =1,
 \end{array}
\right .
$$
where $C_n < D_n$.At this time, we give the noise \{$ N_n$\} of each wireless channel, the maximum delay T of wireless transmission, and the total power $P_{max}$ of the base station.So we can calculate the power consumption $P_n$ required by each user's communication through Shannon's formula .
$$
\begin{array}{cl}
r_n = log_2 ( 1 + \frac{P_n G - L}{N_n}),  \\
T = \frac{d_n}{r_n}.
\end{array}
$$
Where G is the system gain, L is the power lost by transmission.We can make
$$ 
 P_n = \left \{ 
 \begin{array}{rcl}
 0 ,& x_n =-1, \\
 P_{n0} ,& x_n =0, \\
 P_{n1} ,& x_n =1.
 \end{array}
\right.
$$
So we get the optimization problem: $\mathop{max}\limits_{x_n,P_n}  \mathop{\sum}\limits^{n} \rho _n , $ $\mathit{s.t. } \mathop{\sum}\limits^{n} P_n \leq P_{max} $.This is an optimization problem similar to the knapsack problem, and we can obtain the optimal solution of this problem through the dynamic programming algorithm.The recurrence relation is as follows:
$$ 
\mathop{\sum}\limits^{n} \rho _n = F[P][n] = max\left \{ 
 \begin{array}{lcl}
 F[P-P_{n0}][n-1] +  1,&  \\
 F[P-P_{n1}][n-1] + \alpha,&  \\
 F[P][n-1],& 
 \end{array}
\right .
$$
where $n \leq N , P \leq P_{max}$.Then we can get the maximum user total communication quality and its corresponding communication strategy through this recursive formula.

\section{Experiments}
 In this model, the image $X \in \mathbb{C}^{C \times H \times W}$ is first passed through the source encoder $E_s$, which includes one convolution layer and two down-sampling layers, to extract the latent $X_s \in \mathbb{C}^{C_s \times H_s \times W_s}$ from the source image $X$. Here, $C=3$ represents the number of channels, $C_s=256$, $H_s=H/4$, and $W_s=W/4$ denote the dimensions of the extracted feature map.Then $X_s$ passes through the channel encoder $E_c$ and generates $X_c' \in \mathbb{C}^{C_c' \times H_c \times W_c}$, where $C_c'=16$, $H_c=H_s$, and $W_c=W_s$. At the same time, $X_s$ will pass through a policy network to generate a binary mask $W$ to select active features of $X$. The mask $W$ will convert $X_c'$ to $X_c \in \mathbb{C}^{C_c \times L}$, where $C_c=8$ and $L=2 \times H_c \times W_c$. Here we can calculate that the compression ratio of the model in the coding phase is $\mu=\frac{H(X_c)}{H(X)}=1/3$ .
We evaluate the JSCC model with the CIFAR-10 dataset which consists of 50000 training  images with 32 $\times$ 32 pixels. We first train the networks for 150 epochs with a learning rate of  $5 \times 10^{-4}$  and another 150 epochs with a learning rate of  $5 \times 10^{-5}$ . Then, we fix the encoder Es and policy network P , and fine-tune the other modules for another 100 epochs. The batch size is 128.
\subsection{Result based on existing background library}
In this experiment, we tested four sets of images from different scenarios and compared the synthesized images after feature extraction and transmission with the images generated directly through the JSCC model. Under different SNR conditions, we compared the Peak Signal-to-Noise Ratio (PSNR) of images produced by the two methods. The results are shown in Figures~\ref{fig:jscc_result} and~\ref{fig:pm_result}.
\begin{figure}
    \centering
    \includegraphics[width=1\linewidth]{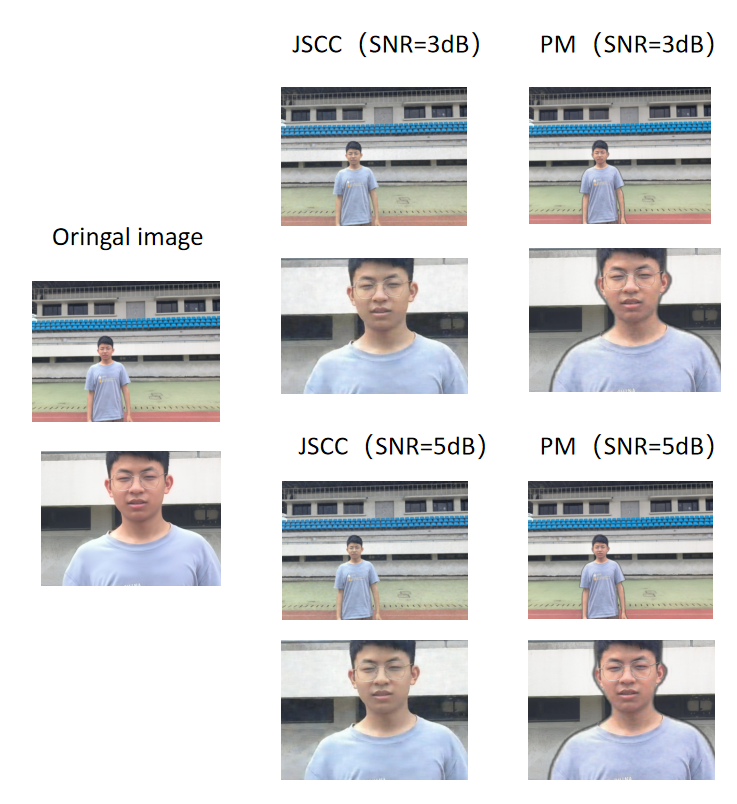}
    \caption{Result of image with JSCC method and PM.}
    \label{fig:enter-label1}
\end{figure}

\begin{figure}
    \centering
    \includegraphics[width=1\linewidth]{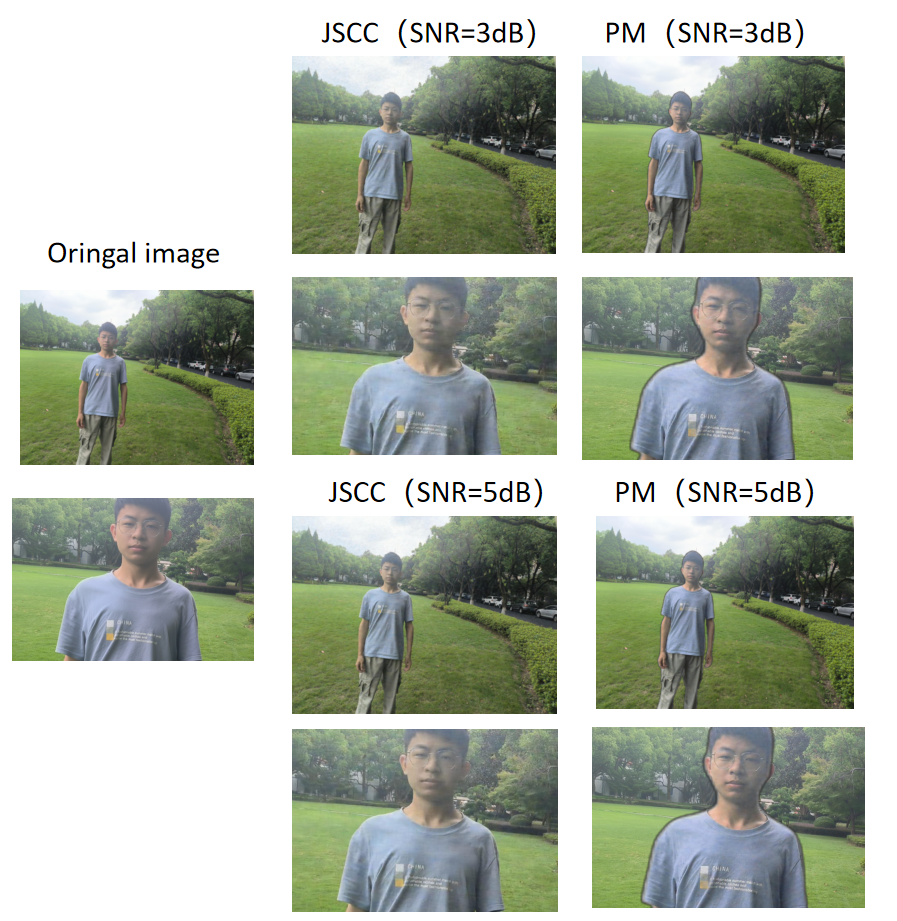}
    \caption{Result of image with JSCC method and PM.}
    \label{fig:enter-label2}
\end{figure}

\begin{figure}[ht]
    \centering
    \includegraphics[width=1\linewidth]{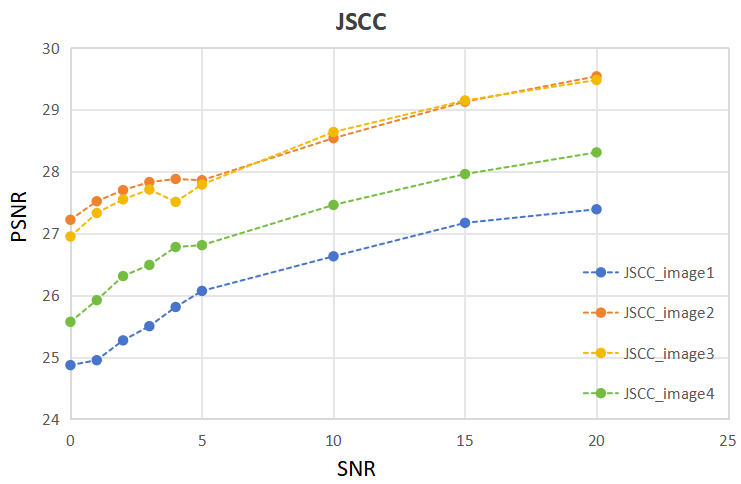}
    \caption{Performance of different images with JSCC method.}
    \label{fig:jscc_result}
\end{figure}

\begin{figure}[ht]
    \centering
    \includegraphics[width=1\linewidth]{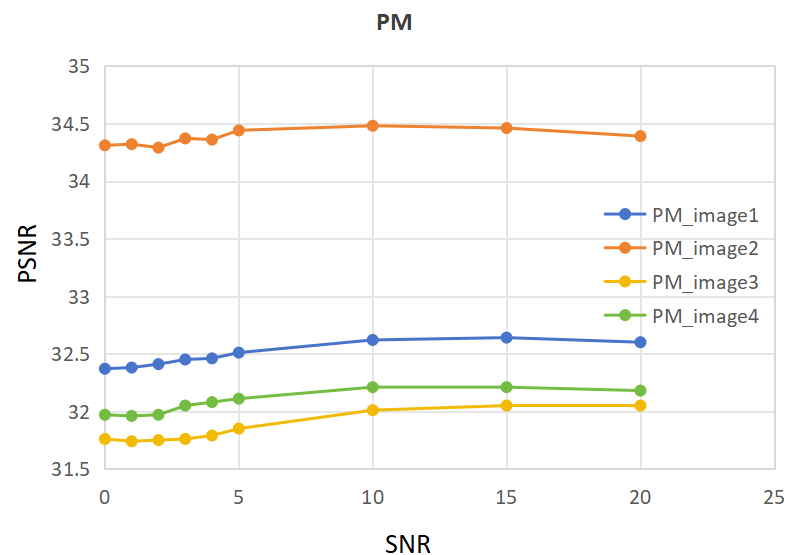}
    \caption{Performance of different images with proposed method.}
    \label{fig:pm_result}
\end{figure}
From the above figure, we can see that the method we mentioned has great advantages over direct transmission through the deep JSCC model. Although the PSNR values of some points on the graph have some anomalies, this is because some features in the image have greater distortion during transmission through the deep JSCC model. But in general, PSNR increases with the increase of SNR.

In addition, compared to the direct JSCC method of transmission, our method also uses the original compression of key information extraction and the coding compression of the coding part of the JSCC model to further reduce the amount of information transmitted while ensuring high-quality transmission, thus reducing the pressure of channel transmission and the power consumption of the transmission base station. This method is applied to application scenarios such as video streaming, surveillance, or remote conferencing, and can compress more than 20 times the amount of information compared with the original data transmission.

\subsection{Result with dynamic background library}
\begin{figure}[ht]
    \centering
    \includegraphics[width=1\linewidth]{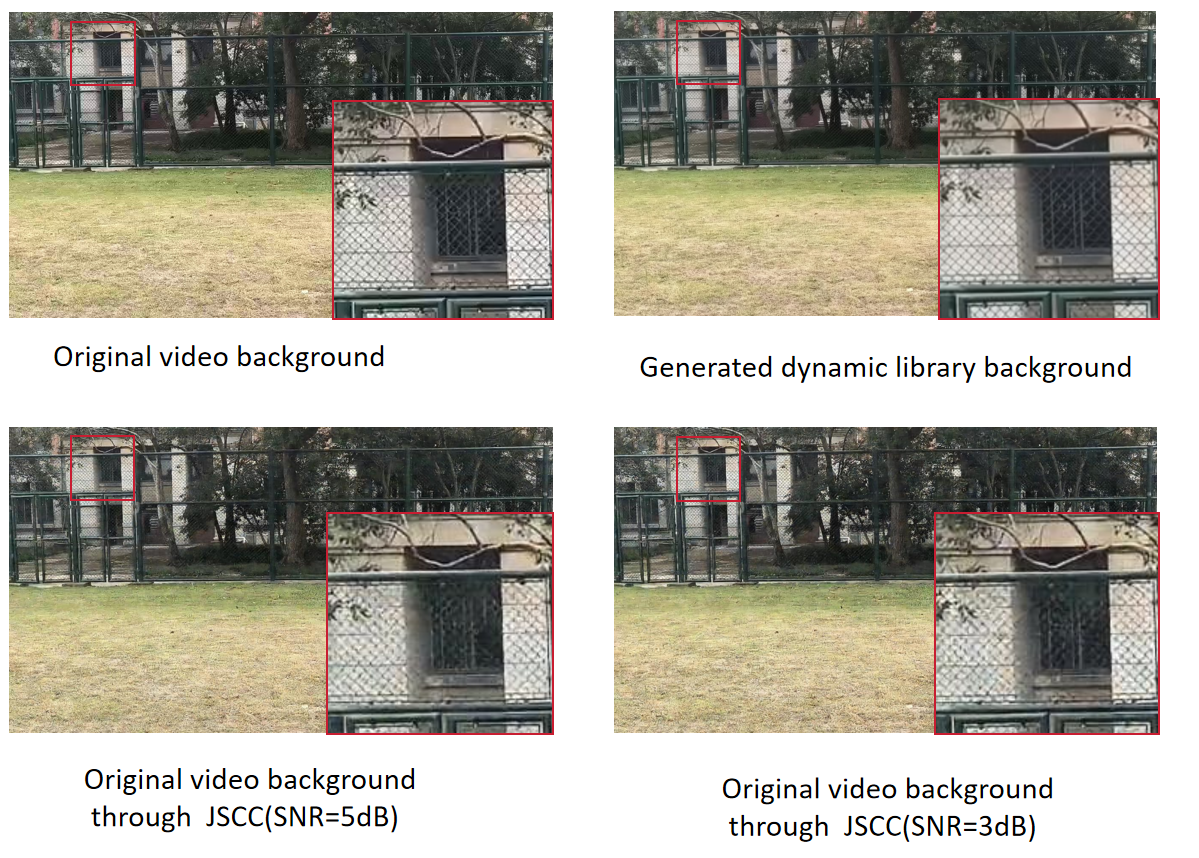}
    \caption{Comparison of results for video backgrounds in different cases.}
    \label{fig:comparison}
\end{figure}
In this experiment, we add the background obtained through video frame splicing as the constructed dynamic background to the background database, and then transmit the video frame by frame directly through the deep JSCC model and the key information extraction method proposed by us, and then compare with the original video frame to obtain PSNR.And in Figure~\ref{fig:comparison} we give the background of the videos in different cases for comparison. As in the previous experiment, we can clearly understand the advantage of our proposed method in image transmission compared to the direct use of JSCC model. So here we only give the background comparison in different cases, by comparing the quality of the background, we can intuitively feel the effect of our proposed method applied to video transmission. Obviously, the quality and sharpness of the background generated by extracting the video information is significantly improved compared to the JSCC model directly.
\begin{figure}
    \centering
    \includegraphics[width=1\linewidth]{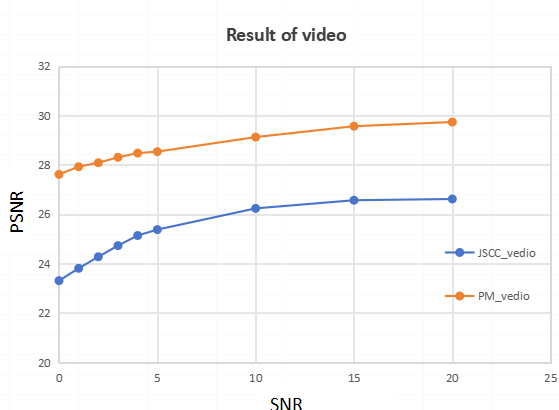}
    \caption{Performance of video with proposed method and JSCC.}
    \label{fig:video_result}
\end{figure}

Based on Figure~\ref{fig:video_result}, it is evident that the results obtained using the dynamic background library construction method produce higher image quality compared to those obtained through direct transmission using the JSCC model. Although the performance gap between the dynamic background construction method and the traditional background library-based transmission is not as large as that observed when comparing with the direct JSCC model transmission, it is still notable. This is because, on the one hand, the dynamic background is constructed using video information, which makes it difficult to maintain the same level of stability across different video frames as in the case of the pre-established background library. On the other hand, the video frames need to be transmitted through the channel and subsequently integrated into the background library at the receiving end, which inevitably incurs some degree of loss during the transmission process. Nevertheless, the dynamic background method still demonstrates a relatively good transmission performance, especially considering that the amount of data transmitted through the channel has been significantly reduced.

\section{Conclusion}
In this paper, we have addressed the significant challenges of high computational complexity and resource demands in wireless image transmission, particularly within the framework of Deep JSCC. We proposed an optimized approach that integrates key information extraction and background matching to enhance both transmission efficiency and image quality. By leveraging advanced feature extraction techniques such as Mediapipe for human detection and Rembg for background removal, our method effectively reduces the amount of data transmitted through the JSCC model. This optimization alleviates computational burden, improves system robustness, and significantly reduces reliance on high-performance GPUs and large video memory, making the system more accessible for resource-constrained devices.

The integration of these techniques into our JSCC model not only addresses the immediate challenges of efficient image transmission but also aligns with the broader goals of 6G networks. The envisioned 6G networks aim to deliver unprecedented levels of performance characterized by ultra-high speeds, minimal latency, and massive connectivity. Our proposed method contributes to these goals by enabling dynamic adaptation to varying channel conditions and user demands, which is essential for the intelligent and adaptive nature of 6G systems. By reducing the computational load and optimizing transmission efficiency, our approach supports the development of self-organizing and self-optimizing networks that can dynamically adjust to changing environments and user requirements.

Our experimental results demonstrate that the proposed method outperforms traditional JSCC transmission in terms of Peak Signal-to-Noise Ratio (PSNR) and overall image quality. The background matching and key information extraction processes significantly mitigate interference from irrelevant background details, allowing for more efficient utilization of available channel resources. Additionally, the resolution detection and adaptive compression strategies ensure that image resolution is dynamically optimized according to available computational resources, making the approach well-suited for practical scenarios with constrained resources. This is particularly important in the context of 6G networks, where the complexity of managing and optimizing such advanced systems requires a more strategic and intelligent approach.

In conclusion, the proposed method provides a promising solution for high-resolution image transmission in wireless communication systems. It effectively balances high image quality with reduced computational demands, making it ideal for applications in mobile communications and multimedia services. The reduced dependence on powerful GPUs and large memory capacity broadens the applicability of the method to more practical, real-world environments. As 6G networks continue to evolve, our research offers a foundation for further exploration into intelligent and adaptive transmission techniques that can meet the stringent requirements of next-generation wireless systems. Future work will focus on leveraging generative AI to address remaining challenges in real-time video transmission and further optimizing existing algorithms to enhance performance in more complex application scenarios, especially under high pressure and limited transmission resources. This will ensure that our approach continues to contribute to the development of high-quality, low-latency image and video transmission solutions that are essential for the success of 6G networks.

\bibliographystyle{IEEEtran}
\bibliography{ref}

\begin{thebibliography}{10}
\providecommand{\url}[1]{#1}
\csname url@samestyle\endcsname
\providecommand{\newblock}{\relax}
\providecommand{\bibinfo}[2]{#2}
\providecommand{\BIBentrySTDinterwordspacing}{\spaceskip=0pt\relax}
\providecommand{\BIBentryALTinterwordstretchfactor}{4}
\providecommand{\BIBentryALTinterwordspacing}{\spaceskip=\fontdimen2\font plus
\BIBentryALTinterwordstretchfactor\fontdimen3\font minus \fontdimen4\font\relax}
\providecommand{\BIBforeignlanguage}[2]{{%
\expandafter\ifx\csname l@#1\endcsname\relax
\typeout{** WARNING: IEEEtran.bst: No hyphenation pattern has been}%
\typeout{** loaded for the language `#1'. Using the pattern for}%
\typeout{** the default language instead.}%
\else
\language=\csname l@#1\endcsname
\fi
#2}}
\providecommand{\BIBdecl}{\relax}
\BIBdecl

\bibitem{ref34}
S.~Dang, O.~Amin, B.~Shihada, and et~al., ``What should 6g be?'' \emph{Nature Electronics}, vol.~3, no.~1, pp. 20--29, 2020.

\bibitem{ref35}
W.~Jiang, B.~Han, M.~A. Habibi, and et~al., ``The road towards 6g: A comprehensive survey,'' \emph{IEEE Open Journal of the Communications Society}, vol.~2, pp. 334--366, 2021.

\bibitem{ref36}
W.~Saad, M.~Bennis, and M.~Chen, ``A vision of 6g wireless systems: Applications, trends, technologies, and open research problems,'' \emph{IEEE network}, vol.~34, no.~3, pp. 134--142, 2019.

\bibitem{10550151}
Z.~Zhao, Z.~Yang, C.~Huang, L.~Wei, Q.~Yang, C.~Zhong, W.~Xu, and Z.~Zhang, ``A joint communication and computation design for distributed {RISs} assisted probabilistic semantic communication in {IIoT},'' \emph{IEEE Internet Things J.}, vol.~11, no.~16, pp. 26\,568--26\,579, Aug. 2024.

\bibitem{ref20}
W.~{Saad}, M.~{Bennis}, and M.~{Chen}, ``A vision of 6g wireless systems: Applications, trends, technologies, and open research problems,'' \emph{IEEE network}, vol.~34, no.~3, pp. 134--142, 2019.

\bibitem{ref21}
N.~{Panwar}, S.~{Sharma}, and A.~K. {Singh}, ``A survey on 5g: The next generation of mobile communication,'' \emph{Physical Communication}, vol.~18, pp. 64--84, 2016.

\bibitem{ref22}
D.~C. {Nguyen}, M.~{Ding}, P.~N. {Pathirana}, A.~{Seneviratne}, J.~{Li}, D.~{Niyato}, O.~{Dobre}, and H.~V. {Poor}, ``6g internet of things: A comprehensive survey,'' \emph{IEEE Internet of Things Journal}, vol.~9, no.~1, pp. 359--383, 2021.

\bibitem{e26050394}
Z.~Zhao, Z.~Yang, M.~Chen, Z.~Zhang, and H.~V. Poor, ``A joint communication and computation design for probabilistic semantic communications,'' \emph{Entropy}, vol.~26, no.~5, Apr. 2024.

\bibitem{ref37}
H.~Viswanathan and P.~E. Mogensen, ``Communications in the 6g era,'' \emph{IEEE access}, vol.~8, pp. 57\,063--57\,074, 2020.

\bibitem{ref38}
H.~Tataria, M.~Shafi, A.~F. Molisch, and et~al., ``6g wireless systems: Vision, requirements, challenges, insights, and opportunities,'' \emph{Proceedings of the IEEE}, vol. 109, no.~7, pp. 1166--1199, 2021.

\bibitem{ZHAO2024107055}
Z.~Zhao, Z.~Yang, X.~Gan, Q.-V. Pham, C.~Huang, W.~Xu, and Z.~Zhang, ``A joint communication and computation design for semantic wireless communication with probability graph,'' \emph{J. Franklin Inst.}, vol. 361, no.~13, p. 107055, Sep. 2024.

\bibitem{ref1}
C.~E. Shannon, ``A mathematical theory of communication,'' \emph{The Bell system technical journal}, vol.~27, no.~3, pp. 379--423, 1948.

\bibitem{ref23}
M.~{Leo}, G.~{Medioni}, M.~{Trivedi}, T.~{Kanade}, and G.~M. {Farinella}, ``Computer vision for assistive technologies,'' \emph{Computer Vision and Image Understanding}, vol. 154, pp. 1--15, 2017.

\bibitem{ref24}
W.~{Yang}, H.~{Du}, Z.~Q. {Liew}, W.~Y.~B. {Lim}, Z.~{Xiong}, D.~{Niyato}, X.~{Chi}, X.~{Shen}, and C.~{Miao}, ``Semantic communications for future internet: Fundamentals, applications, and challenges,'' \emph{IEEE Communications Surveys \& Tutorials}, vol.~25, no.~1, pp. 213--250, 2022.

\bibitem{ref25}
Y.~{Tian}, G.~{Pan}, and M.~S. {Alouini}, ``Applying deep-learning-based computer vision to wireless communications: Methodologies, opportunities, and challenges,'' \emph{IEEE Open Journal of the Communications Society}, vol.~2, pp. 132--143, 2020.

\bibitem{ref26}
X.~Luo, H.~H. Chen, and Q.~Guo, ``Semantic communications: Overview, open issues, and future research directions,'' \emph{IEEE Wireless Communications}, vol.~29, no.~1, pp. 210--219, 2022.

\bibitem{ref27}
J.~{Bao}, P.~{Basu}, M.~{Dean}, C.~{Partridge}, A.~{Swami}, W.~{Leland}, and J.~A. {Hendler}, ``Towards a theory of semantic communication,'' in \emph{2011 IEEE Network Science Workshop}, 2011, pp. 110--117.

\bibitem{ref28}
P.~{Basu}, J.~{Bao}, M.~{Dean}, and J.~{Hendler}, ``Preserving quality of information by using semantic relationships,'' \emph{Pervasive and Mobile Computing}, vol.~11, pp. 188--202, 2014.

\bibitem{10915662}
Z.~Zhao, Z.~Yang, Y.~Hu, C.~Zhu, M.~Shikh-Bahaei, W.~Xu, Z.~Zhang, and K.~Huang, ``Compression ratio allocation for probabilistic semantic communication with {RSMA},'' \emph{IEEE Trans. Commun.}, pp. 1--1, 2025.

\bibitem{ref29}
J.~Park and S.~W. Yoon, ``Transmit what you need: task-adaptive semantic communications for visual information,'' \emph{arXiv preprint arXiv:2412.13646}, 2024.

\bibitem{ref30}
J.~Y. Zhu, T.~Park, P.~Isola, and et~al., ``Unpaired image-to-image translation using cycle-consistent adversarial networks,'' in \emph{Proceedings of the IEEE international conference on computer vision}, 2017, pp. 2223--2232.

\bibitem{ref31}
E.~Grassucci, J.~Park, S.~Barbarossa, and et~al., ``Generative ai meets semantic communication: Evolution and revolution of communication tasks,'' \emph{arXiv preprint arXiv:2401.06803}, 2024.

\bibitem{ref32}
L.~Xia, Y.~Sun, C.~Liang, and et~al., ``Generative ai for semantic communication: Architecture, challenges, and outlook,'' \emph{IEEE Wireless Communications}, vol.~32, no.~1, pp. 132--140, 2025.

\bibitem{ref33}
C.~Liang, H.~Du, Y.~Sun, and et~al., ``Generative ai-driven semantic communication networks: Architecture, technologies and applications,'' \emph{IEEE Transactions on Cognitive Communications and Networking}, 2024.

\bibitem{ref2}
A.~{Vosoughi}, P.~C. {Cosman}, and L.~B. {Milstein}, ``Joint source-channel coding and unequal error protection for video plus depth,'' \emph{IEEE Signal Processing Letters}, vol.~22, no.~1, pp. 31--34, 2014.

\bibitem{ref3}
S.~{Heinen} and P.~{Vary}, ``Transactions papers source-optimized channel coding for digital transmission channels,'' \emph{IEEE transactions on communications}, vol.~53, no.~4, pp. 592--600, 2005.

\bibitem{ref4}
V.~{Bozantizis} and F.~{Ali}, ``Combined vector quantisation and index assignment with embedded redundancy for noisy channels,'' \emph{Electronics Letters}, vol.~36, no.~20, pp. 1711--1713, 2000.

\bibitem{ref48}
A.~Shirazinia, S.~Chatterjee, and M.~Skoglund, ``Joint source-channel vector quantization for compressed sensing,'' \emph{IEEE transactions on signal processing}, vol.~62, no.~14, pp. 3667--3681, 2014.

\bibitem{ref49}
M.~Fresia and G.~Caire, ``A linear encoding approach to index assignment in lossy source-channel coding,'' \emph{IEEE transactions on information theory}, vol.~56, no.~3, pp. 1322--1344, 2010.

\bibitem{ref50}
O.~Fresnedo, J.~P. Gonzalez-Coma, M.~Hassanin, and et~al., ``Evaluation of analog joint source-channel coding systems for multiple access channels,'' \emph{IEEE Transactions on Communications}, vol.~63, no.~6, pp. 2312--2324, 2015.

\bibitem{10734747}
Z.~Zhao, Z.~Yang, Q.~Yang, C.~Huang, M.~Shikh-Bahaei, and Z.~Zhang, ``Sum rate maximization for distributed riss assisted probabilistic semantic communication,'' in \emph{2024 IEEE 34th International Workshop on Machine Learning for Signal Processing (MLSP)}, 2024, pp. 1--6.

\bibitem{ref39}
S.~Dörner, S.~Cammerer, J.~Hoydis, and et~al., ``Deep learning based communication over the air,'' \emph{IEEE Journal of Selected Topics in Signal Processing}, vol.~12, no.~1, pp. 132--143, 2017.

\bibitem{ref40}
W.~Jiang, ``Graph-based deep learning for communication networks: A survey,'' \emph{Computer Communications}, vol. 185, pp. 40--54, 2022.

\bibitem{ref41}
H.~Kim, Y.~Jiang, R.~Rana, and et~al., ``Communication algorithms via deep learning,'' \emph{arXiv preprint arXiv:1805.09317}, 2018.

\bibitem{ref42}
L.~Dai, R.~Jiao, F.~Adachi, and et~al., ``Deep learning for wireless communications: An emerging interdisciplinary paradigm,'' \emph{IEEE Wireless Communications}, vol.~27, no.~4, pp. 133--139, 2020.

\bibitem{ref43}
H.~Xie, Z.~Qin, G.~Y. Li, and et~al., ``Deep learning enabled semantic communication systems,'' \emph{IEEE transactions on signal processing}, vol.~69, pp. 2663--2675, 2021.

\bibitem{ref44}
H.~Ye, L.~Liang, G.~Y. Li, and et~al., ``Deep learning-based end-to-end wireless communication systems with conditional gans as unknown channels,'' \emph{IEEE Transactions on Wireless Communications}, vol.~19, no.~5, pp. 3133--3143, 2020.

\bibitem{ref45}
T.~Wang, C.~K. Wen, H.~Wang, and et~al., ``Deep learning for wireless physical layer: Opportunities and challenges,'' \emph{China Communications}, vol.~14, no.~11, pp. 92--111, 2017.

\bibitem{ref46}
H.~Huang, S.~Guo, G.~Gui, and et~al., ``Deep learning for physical-layer 5g wireless techniques: Opportunities, challenges and solutions,'' \emph{IEEE Wireless Communications}, vol.~27, no.~1, pp. 214--222, 2019.

\bibitem{ref47}
Z.~Tang, S.~Shi, W.~Wang, and et~al., ``Communication-efficient distributed deep learning: A comprehensive survey,'' \emph{arXiv preprint arXiv:2003.06307}, 2020.

\bibitem{11006980}
Z.~Zhao, Z.~Yang, M.~Chen, C.~Zhu, W.~Xu, Z.~Zhang, and K.~Huang, ``Energy-efficient probabilistic semantic communication over space-air-ground integrated networks,'' \emph{IEEE Trans. Wireless Commun.}, pp. 1--1, 2025.

\bibitem{ref5}
E.~Bourtsoulatze, D.~B. Kurka, and D.~Gündüz, ``Deep joint source-channel coding for wireless image transmission,'' \emph{IEEE Transactions on Cognitive Communications and Networking}, vol.~5, no.~3, pp. 567--579, 2019.

\bibitem{ref6}
D.~B. {Kurka} and D.~{Gündüz}, ``Successive refinement of images with deep joint source-channel coding,'' in \emph{2019 IEEE 20th International Workshop on Signal Processing Advances in Wireless Communications (SPAWC)}, 2019, pp. 1--5.

\bibitem{ref7}
------, ``Deepjscc-f: Deep joint source-channel coding of images with feedback,'' \emph{IEEE Journal on Selected Areas in Information Theory}, 2020.

\bibitem{ref8}
J.~{Xu}, B.~{Ai}, W.~{Chen}, A.~{Yang}, P.~{Sun}, and M.~{Rodrigues}, ``Wireless image transmission using deep source channel coding with attention modules,'' \emph{IEEE Transactions on Circuits and Systems for Video Technology}, 2021.

\bibitem{ref9}
M.~Song, N.~Ma, C.~Dong, and et~al., ``Deep joint source-channel coding for wireless image transmission with adaptive models,'' \emph{Electronics}, vol.~12, no.~22, p. 4637, 2023.

\bibitem{ref10}
W.~Chen, Y.~Chen, Q.~Yang, and et~al., ``Deep joint source-channel coding for wireless image transmission with entropy-aware adaptive rate control,'' in \emph{GLOBECOM 2023-2023 IEEE Global Communications Conference}.\hskip 1em plus 0.5em minus 0.4em\relax IEEE, 2023, pp. 2239--2244.

\bibitem{ref11}
M.~A. Jarrahi, E.~Bourtsoulatze, and V.~Abolghasemi, ``Joint source-channel coding for wireless image transmission: A deep compressed-sensing based method,'' in \emph{2024 IEEE Wireless Communications and Networking Conference (WCNC)}.\hskip 1em plus 0.5em minus 0.4em\relax IEEE, 2024, pp. 1--6.

\bibitem{ref12}
K.~{Choi}, K.~{Tatwawadi}, A.~{Grover}, T.~{Weissman}, and S.~{Ermon}, ``Neural joint sourcechannel coding,'' in \emph{International Conference on Machine Learning}.\hskip 1em plus 0.5em minus 0.4em\relax PMLR, 2019, pp. 1182--1192.

\bibitem{ref13}
M.~Yang and H.~S. Kim, ``Deep joint source-channel coding for wireless image transmission with adaptive rate control,'' in \emph{ICASSP 2022-2022 IEEE International Conference on Acoustics, Speech and Signal Processing (ICASSP)}.\hskip 1em plus 0.5em minus 0.4em\relax IEEE, 2022, pp. 5193--5197.

\bibitem{ref14}
R.~Yamamoto, Y.~Inoue, and D.~Hisano, ``Deep joint source-channel coding using overlap image division for block noise reduction,'' in \emph{2024 IEEE 99th Vehicular Technology Conference (VTC2024-Spring)}.\hskip 1em plus 0.5em minus 0.4em\relax IEEE, 2024, pp. 1--6.

\bibitem{ref15}
M.~Naseri, P.~Ashtari, M.~Seif, and et~al., ``Deep learning-based image compression for wireless communications: Impacts on reliability, throughput, and latency,'' \emph{arXiv preprint arXiv:2411.10650}, 2024.

\bibitem{ref16}
C.~Lugaresi, J.~Tang, H.~Nash, and et~al., ``Mediapipe: A framework for building perception pipelines,'' \emph{arXiv preprint arXiv:1906.08172}, 2019.

\bibitem{ref17}
------, ``Mediapipe: A framework for perceiving and processing reality,'' in \emph{Third workshop on computer vision for AR/VR at IEEE computer vision and pattern recognition (CVPR)}, 2019.

\bibitem{ref18}
E.~Rublee, V.~Rabaud, K.~Konolige, and et~al., ``Orb: An efficient alternative to sift or surf,'' in \emph{2011 International conference on computer vision}.\hskip 1em plus 0.5em minus 0.4em\relax Ieee, 2011, pp. 2564--2571.

\bibitem{ref19}
Z.~Huang, T.~Zhang, W.~Heng, and et~al., ``Real-time intermediate flow estimation for video frame interpolation,'' in \emph{European Conference on Computer Vision}.\hskip 1em plus 0.5em minus 0.4em\relax Cham: Springer Nature Switzerland, 2022, pp. 624--642.

\end{thebibliography}

\end{document}